\documentclass{article}
\usepackage{spconf,amsmath,graphicx}
\usepackage{algorithm}
\usepackage{algorithmic}
\usepackage{amsfonts}
\usepackage{amsthm}
\usepackage{amsmath}
\usepackage{amscd}
\usepackage[latin2]{inputenc}
\usepackage{t1enc}
\usepackage[mathscr]{eucal}
\usepackage{indentfirst}
\usepackage{graphicx}
\usepackage{graphics}
\usepackage{subfig}
\usepackage{pict2e}
\usepackage{epic}
\usepackage{xcolor,colortbl}
\numberwithin{equation}{section}
\usepackage{epstopdf}

\newcommand{\dn}[1]{{\bfseries{\textcolor{purple}{#1}}}}
\title{Clustering of Nonnegative Data and an Application to Matrix Completion}
\name{C. Strohmeier, D. Needell\thanks{Partially supported by NSF CAREER
$\#$1348721, and NSF BIGDATA $\#$1740325.}}
\address{Department of Mathematics, University of California, Los Angeles}
\begin{document}
%\ninept
%
\maketitle
\begin{abstract}
In this paper, we propose a simple algorithm to cluster nonnegative data lying in disjoint subspaces. We analyze its performance in relation to a certain measure of correlation between said subspaces. We use our clustering algorithm to develop a matrix completion algorithm which can outperform standard matrix completion algorithms on data matrices satisfying certain natural conditions. 
\end{abstract}

\section{Introduction} 

Real world data is often high dimensional, that is to say a given data point may be modeled by a vector in a high dimensional Euclidean space. While genuinely high dimensional data would be prohibitively difficult to analyze, most data encountered are effectively low dimensional. For instance, the data points may approximately lie in some (a priori unknown) low dimensional subspace of their ambient space. Moreover, one often encounters data which is nonnegative in the sense that each of its entries is nonnegative, such as is data stemming from user surveys, rating systems, or biomedical monitoring. Here, we consider such low dimensional data and address the problem of data completion via two other data oriented tasks, clustering and nonnegative factorization.

An important problem in data science is that of data completion, and matrix completion in particular \cite{RefWorks:390}. Namely, one wishes to recover an unknown matrix from only a subset of its entries. To make the problem well-posed, one typically assumes that the underlying matrix is also low-rank, a reasonable assumption in many applications where there is a small number of intrinsic features that describe the large-scale data.  One typically also assumes that the observed entries are selected in a ``nice" way, such as uniformly at random, to avoid degenerate sampling patterns. There are now many provably robust methods to matrix completion including convex optimization programs \cite{RefWorks:32,RefWorks:375} and projection based methods \cite{RefWorks:698,keshavan2010matrix}. 

\textit{Clustering} is another typical problem in data science whose aim is to cluster, or group, unlabeled data. That is, one has a data set consisting of two or more families of data points such that members of each family share intrinsic characteristics. Based on these intrinsic characteristics, one must sort the data into its different families. There are now many methods to cluster data along with a wide array of theoretical and empirical support, see e.g. \cite{jain1999data,xu2005survey,alelyani2018feature} and references therein. Although there are now many sophisticated methods for clustering, the simpler $k$-means clustering method \cite{david2003chapter}, which aims to separate the data points into $k$ clusters so that each point belongs to a cluster with the nearest mean, is still useful in many applications. However, like many others, $k$-means fails in most applications where the data families live in some low dimensional subspaces, where linear separability need not be apparent. 

Lastly, another useful tool we will utilize is nonnegative matrix factorization (NMF) \cite{tandon2010sparse,berry2007algorithms}. Concretely, the problem of NMF is to factor an $M\times N$ data matrix $X$ into $X \approx AS$ where $A$ is a non-negative $M\times T$ matrix and $S$ a non-negative $T\times N$ matrix. The parameter $T$ corresponds to the number of \textit{topics} to represent the data, the matrix $A$ then gives a topic representation for each of the $M$ variables, and $S$ a topic representation for each of the $N$ users (for example). Concretely, we can view $A_{ij}$ as an indicator of how important the $i$th variable is for the $j$th topic, and $S_{jk}$ as how important the $j$th topic is to user $k$. This structure implicitly reveals topics in the data, which can be interpreted on their own or used as features in other data processing tasks. This type of NMF is called \textit{unsupervised} representation since it works only on the raw data, without any other observation information. NMF is also by now a standard tool in dimensionality reduction. The advantage of this factorization compared to other dimensionality reduction techniques such as PCA, IMF, etc. is that the nonnegativity constraints enforce a certain locality, and hence interpretability, of its hidden features. See for instance \cite{lee2001algorithms}, \cite{lee1999learning}.

In this paper, we discuss the clustering and completion problems in the context of nonnegative data belonging to low dimensional subspaces of a high dimensional Euclidean space. This additional structure appears in data arising from an abundance of applications, ranging from collaborative filtering to multi-class learning, where the structure may arise from subgroups of e.g. users that have similar preferences. Utilizing this additional structure allows us to decrease the number of observations needed and/or decrease the reconstruction error. In particular, we present a clustering algorithm based on nonnegative matrix factorization (NMF), Algorithm \ref{alg1}, and examine the relationship between its performance and a certain measure of correlation between the two families of data points that we are trying to separate (cluster). We then discuss an application to a certain "block completion" algorithm, which can yield significant improvements compared to standard matrix completion algorithms regarding recovery error for matrices satisfying a very natural low-rank type condition.

\section{Clustering Via NMF}

Consider a data set consisting of vectors $x_i$ belonging to one of two disjoint low dimensional subspaces $W_1, W_2 \leq \mathbb{R}^n$. Suppose further that the entries of the data vectors are all nonnegative. The problem is to sort the data according to its respective subspace. While standard clustering techniques applied to the data fail in general, one may take advantage of a certain orthogonalizing effect of NMF. 
To explain the first step in our proposed simple method, let us interpret our $m$ data points $x_i\in\mathbb{R}^n$ as row vectors and concatenate them into the data matrix $X\in\mathbb{R}^{m\times n}$. 
Suppose we have an upper bound for the sum of the dimensions of the subspaces:
$$\text{dim}W_1 + \text{dim}W_2 \leq r$$
We perform NMF with $r$ hidden features (topics) to factor $X$ into a product of a weight matrix $W\in\mathbb{R}^{m\times r}$ and a hidden feature matrix $H\in\mathbb{R}^{r\times n}$. We then cluster not on the original data, but on the rows of the weight matrix $W$. This is described succinctly as Algorithm \ref{alg1}.

\begin{algorithm}
\caption{Clustering Via NMF}
\label{alg1}
\begin{algorithmic}

\STATE  1. Input: Data matrix X, upper bound for rank r.
\STATE 2. NMF: Perform NMF with r hidden features to write X = WH
\STATE 3. Apply k-means to rows of W

\end{algorithmic}
\end{algorithm}

Algorithm \ref{alg1} proves to be very effective, even in the presence of considerable noise. One explanation for this phenomenon is that NMF exhibits a certain orthogonalization effect. In particular, it is not necessarily true that if one forms a nonnegative matrix $H'$ whose rows are a given set of $r$ linearly independent nonnegative vectors belonging to $W_1 + W_2$ that one may still be able to factor $X = W'H'$ for some \textit{nonnegative} $W'$. This prevents a degree of "mixing" of hidden feature basis vectors associated to the two subspaces when performing NMF. The result is that the rows of any weight matrix $W$ obtained via NMF applied to our data matrix will have very small values at entries corresponding to basis vectors for a family in which is does not belong. Put simply, NMF has reduced the problem of clustering nonnegative data belonging to general low dimensional subspaces to that of clustering nonnegative data belonging to low dimensional \textit{coordinate} subspaces (thus NMF has "orthogonalized" the original subspaces). 

Below we illustrate the relationship between success of clustering and a certain measure of correlation between subspaces. 

Suppose $U, V \subset \mathbb{R}^n$ are two subspaces of the same dimension $r$. We define the correlation measure $\alpha(U, V)$ between them as 
$$\alpha(U, V) = \frac{1}{r}tr(P_U P_V),$$
where $P_U$ denotes the orthogonal projection onto $U$ and similarly for $P_V$. Note that for any two subspaces we have $0 \leq \alpha(U, V) \leq 1$, $\alpha(U, V) = 0$ iff $U$ and $V$ are orthogonal and $\alpha(U, V) = 1$ iff $U = V$. 

Naturally, one expects that as the correlation measure between subspaces increases, they become more difficult to cluster. This is indeed the case.

Fix $n, r$ with $r \ll n$ and consider an $r \times n$ random matrix with i.i.d. entries uniformly distributed between 0 and 1. A particular instance of this random matrix is a proxy for an $r$-dimensional subspace of $\mathbb{R}^n$. Abusing notation, we will refer to such an instance as $U$. We construct matrices $V$ with various correlation measures through a multiplication by a matrix of the form $\text{exp}(tA)$ where $t$ is a scalar and $A$ is a random skew-symmetric matrix (obtained by skew-symmetrizing an instance of a random standard Gaussian matrix). Thus $t = 0$ implies $U = V$, and larger $t$ implies a greater correlation measure between the subspaces. Of course $t$ cannot be too large, otherwise $V$ may possess negative entries. 

We will consider various pairs of $U$ and $V$ as constructed above. Let $U$-block be the product of an instance of an $m \times r$ standard uniform matrix with $U$, and construct $V$-block similarly. We concatenate these two row matrices to obtain our $2m \times n$ row matrix $X$. 

\begin{figure}
\includegraphics[width=8cm]{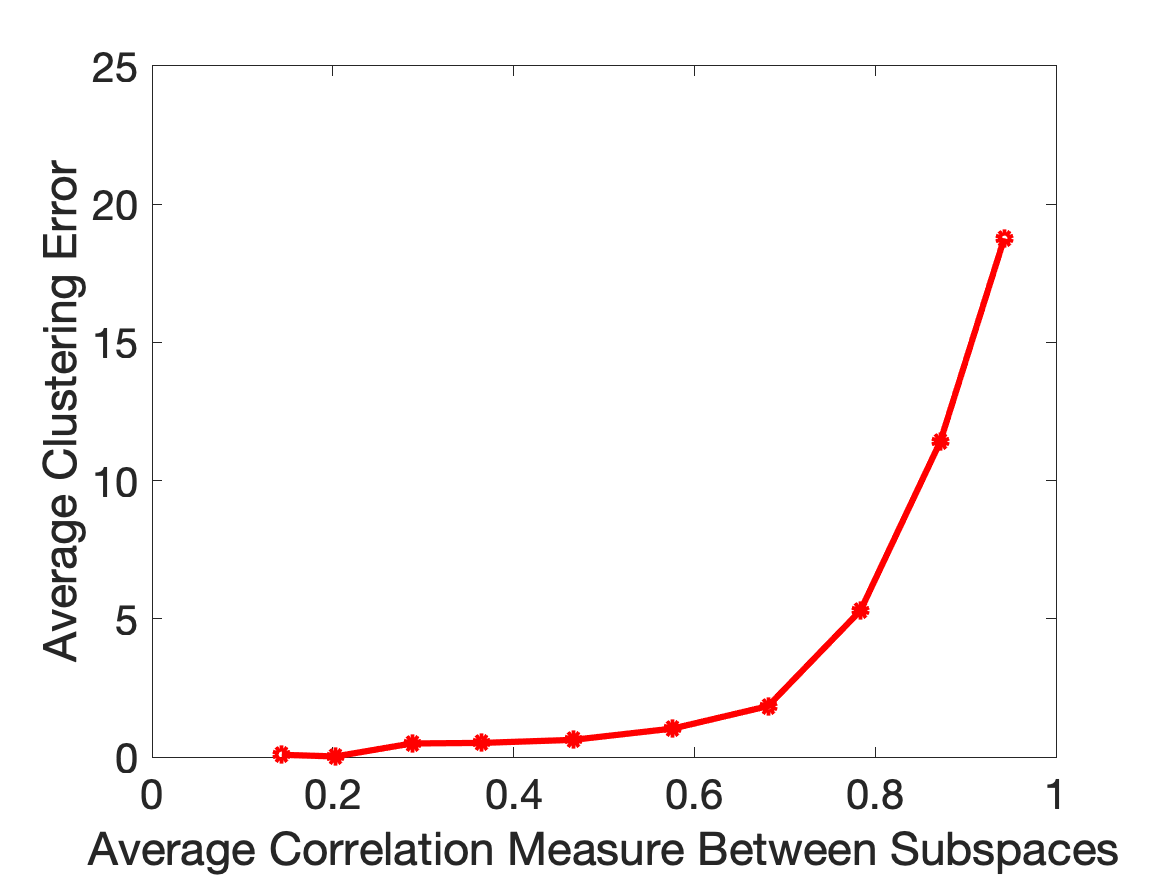}
\caption{Average clustering error (average number of misclassified data points over $100$ trials) as a function of the average correlation measure $\alpha(U, V) = \frac{1}{r}tr(P_U P_V)$ between the subspaces $U, V$ which generated the data.}\label{fig1}
\end{figure}

Figure \ref{fig1} represents the average clustering error under the above model with $m = 100, n = 80, r = 5$. Each simulation consists of 100 samples. As noted, the accuracy is smaller for smaller values of $\alpha$, when the subspaces are closer to being orthogonal.

%\begin{center}
%\begin{tabular}{| c | c | c | c | c | c |} 
%
%\hline
% \multicolumn{6}{|c|}{Average Clustering Error} \\
%\hline
%
%\hline
%Average Angle & $\alpha = .36$ & $\alpha = .57$ & $\alpha = .68 $ & $\alpha = .79$ & $\alpha  = .87$ \\
%\hline
%Average Error & 0 & 1 & 2 & 4 & 10 \\
%\hline
%
%
%
%\end{tabular}
%\end{center}

\section{Application to Matrix Completion}
In many applications where one wishes to perform data completion, the data points naturally lie in different subspaces. For example, in collaborative filtering where the data represents users and their ratings of certain products, the overall data matrix may be low rank since there are a few underlying features that describe the users or products (e.g. movie genre, user demographics or preferences, etc.). However, it may more often be the case that even more structure is present, for example the users may be divided into similar blocks so that each block is of even lower rank than the overall matrix. If such blocks can be identified, performing matrix completion on each block individually will result in lower error and/or the need for fewer measurements.
 
Here we thus consider the problem of matrix completion, applied to matrices of the form considered in the previous section (namely concatenations of blocks). In fact we allow for matrices obtained by permuting the rows of such matrices, since the data is unlabeled. While one can attempt to apply standard matrix completion, we demonstrate that improvements can be made if one exploits the additional structure of our data matrices. 

We introduce a novel matrix completion algorithm, Algorithm \ref{alg2}, which we refer to as Block Completion. The algorithm proceeds by first applying an initial, "basic" matrix completion to the entire data matrix.

In practice, applying a finite number of iterates of a standard matrix completion algorithm may yield matrices which have negative entries, even if all of the entries of the original matrix are nonnegative. We avoid this problem by using a certain simultaneous matrix completion and nonnegative matrix factorization algorithm, MC-NMF as in \cite{MCNMF12}, to implement our basic completion. MC-NMF takes in an incomplete, low rank, nonnegative matrix $X$, a set of sampled values, and an estimate for its rank, and at each iterate returns a completed pair of low rank nonnegative factors $W, H$ such that after sufficiently many iterates one obtains $X \approx WH$. In short, the "basic completion" algorithm used in this paper simply applies MC-NMF with a fixed number of iterates and returns the product $WH$ of its outputs. 

Although the error from this initial matrix completion may be large, empirically one often finds that Algorithm \ref{alg1} nonetheless succeeds in clustering the (noisy) data points. We obtain data sub-matrices (blocks) from these clusters, and apply basic completion to each block using the original observed entries. The point is that the blocks are lower rank, and so one expects an improvement in recovery error. We concatenate the completed blocks in the obvious way and take this as the output for block completion.

\begin{algorithm}
\caption{Block Completion}
\label{alg2}
\begin{algorithmic}

\STATE 1. Initialization:  Mask $\Omega$ of observed entries of data matrix $X$, upper bound for rank $r$.
\STATE 2. Basic Completion: Apply standard matrix completion algorithm to whole matrix X using observed entries $\Omega$
\STATE 3. Sort Blocks: Apply clustering via NMF (Algorithm \ref{alg1}) to obtain matrices $A$ and $B$ by concatenating the data vectors belonging to respective clusters.
\STATE 4. Complete Blocks: Using masks derived from original mask, apply basic completion to both $A$ and $B$ individually.
\STATE 5. Reassemble: construct the full completed matrix from the completed blocks. 

\end{algorithmic}
\end{algorithm}

\begin{figure}
\includegraphics[width=8cm]{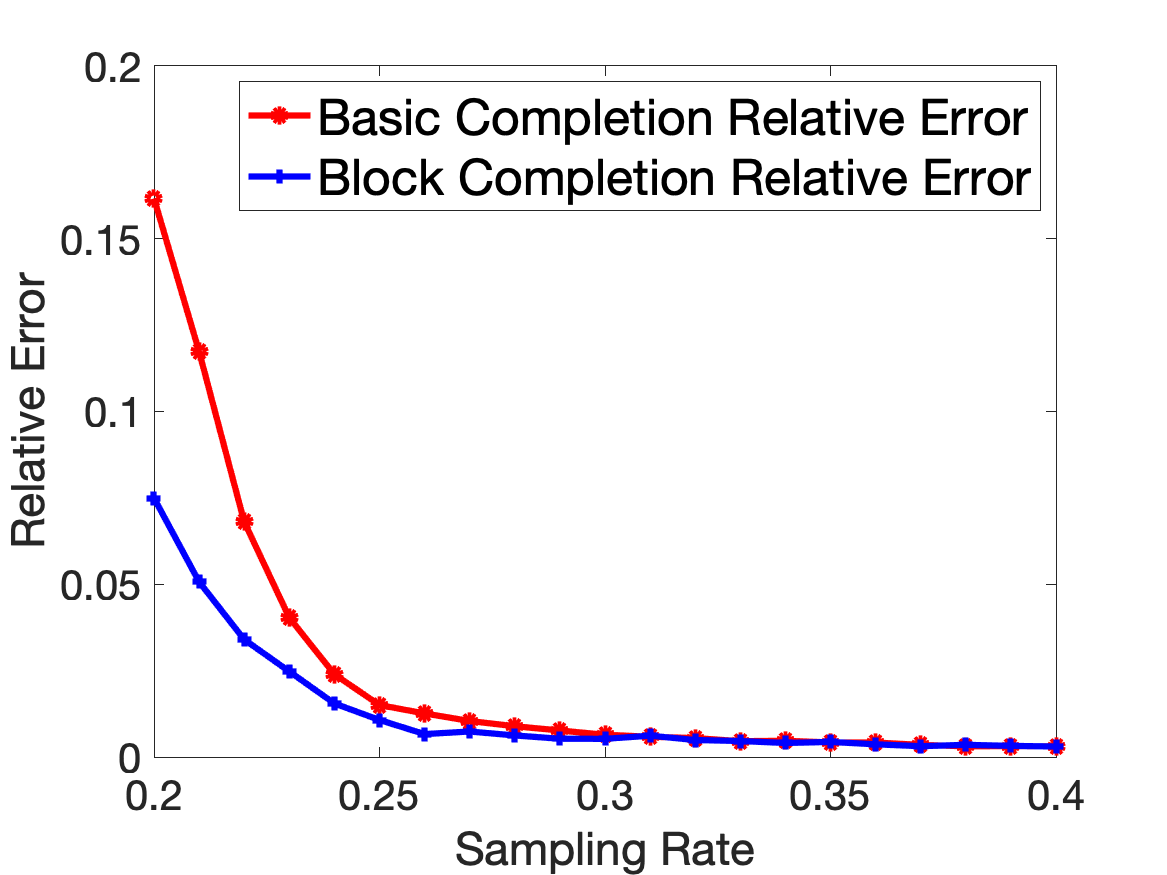}
\caption{Comparison of Relative Error Between Completion Algorithms as a function of the sampling rate $p$.}\label{fig2}
\end{figure}

To test this approach, we recorded a comparison of standard completion with block completion. Our data matrices were constructed by concatenating two low rank blocks. Each block was the product of an instance of a standard uniform $100 \times 5$ matrix with a standard uniform $5 \times 80$ matrix. The observed entries corresponded to a Bernoulli matrix with various sampling rates $p$. We used 500 iterates of the basic completion algorithm described above. The result represents the average relative error of one hundred random samples. Figure \ref{fig2} demonstrates the advantages of such an approach. Indeed, for low sampling rates the relative error when applying block completion is significantly lower than that of basic completion. Figure \ref{fig3} shows an example of the obtained results.

\iffalse

\begin{figure}
\centering

\includegraphics[height=5cm,width=3cm]{a1.png}\\

\includegraphics[height=5cm,width=3cm]{a2.png}\\
\includegraphics[height=5cm,width=3cm]{a3.png}
\caption{Example of original matrix (top), reconstructed matrix via standard matrix completion (middle), and by the proposed block completion method (bottom). Errors for standard and proposed approach are \dn{fill in}, respectively. }\label{fig3}
\end{figure}

\fi

\iffalse

\begin{figure}

\subfloat{
\includegraphics{a1.png}
}
\subfiloat{
\includegraphics{a2.png}
}

\subfloat{
\includegraphics{a3.png}
}

\caption{Example of original matrix (top left), reconstructed matrix via standard matrix completion (top right), and by the proposed block completion method (bottom left). Errors for standard and proposed approach are .17 and .05, respectively. }\label{fig3}
\label{fig:whatever}
\end{figure}

\fi

\begin{figure}

\centering

\includegraphics[height = 6cm , width = 9cm ]{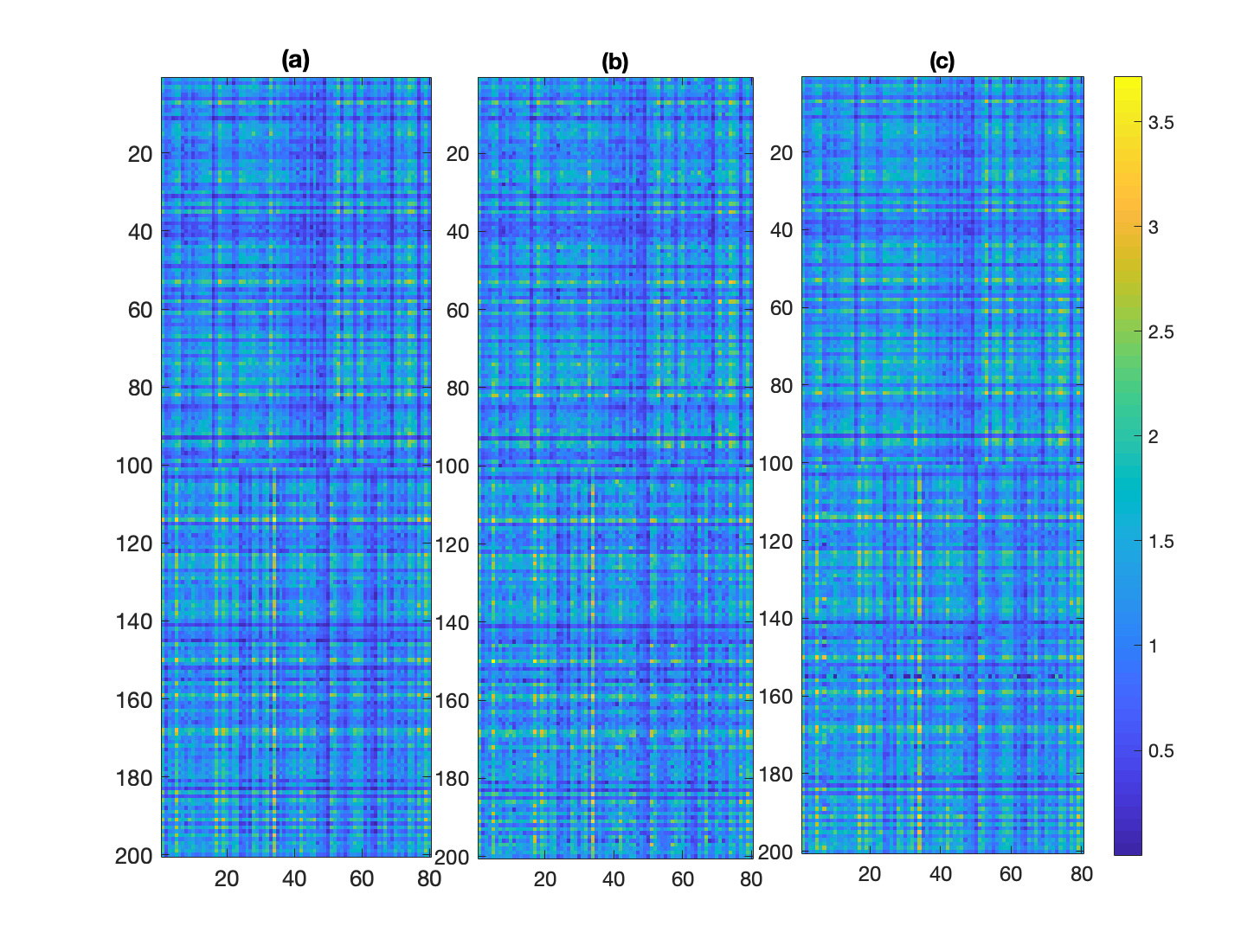}

\caption{Example of original matrix (a), reconstructed matrix via standard matrix completion (b), and by the proposed block completion method (c). Errors for standard and proposed approach are .17 and .05, respectively. }\label{fig3}
\label{fig:whatever}

\end{figure}

\section{Conclusion}

We have presented a simple clustering algorithm based on NMF which is very effective in separating nonnegative data belonging to disjoint, low-dimensional subspaces, even with considerable noise. As an application, we introduced a matrix completion algorithm designed to more accurately complete certain matrices satisfying a low-rank type condition. This algorithm has the potential to yield non-trivial insights from incomplete data matrices comprised of data points belonging to two or more different families.

%\begin{center}
%\begin{tabular}{| c | c | c | c |} 
%
%\hline
% \multicolumn{4}{|c|}{Comparison of Relative Error Between Completion Algorithms} \\
%\hline
%
%\hline
%Sampling Rate & p = .3 & p = .25 & p = .2 \\
%\hline
%Standard Completion & .008 & .017 & .16 \\
%\hline
%Block Completion & .005 & .009 & .08
%\\
%\hline
%
%
%\end{tabular}
%\end{center}

\bibliographystyle{myalpha}
\bibliography{strings,bib}

\newcommand{\etalchar}[1]{$^{#1}$}
\begin{thebibliography}{XYWZ12}

\bibitem[ATL18]{alelyani2018feature}
S.~Alelyani, J.~Tang, and H.~Liu.
\newblock Feature selection for clustering: A review.
\newblock In {\em Data Clustering}, pages 29--60. Chapman and Hall/CRC, 2018.

\bibitem[BBL{\etalchar{+}}07]{berry2007algorithms}
M.~W. Berry, M.~Browne, A.~N. Langville, V.~P. Pauca, and R.~J. Plemmons.
\newblock Algorithms and applications for approximate nonnegative matrix
  factorization.
\newblock {\em Computational statistics \& data analysis}, 52(1):155--173,
  2007.

\bibitem[CP10]{RefWorks:375}
E.~J. Cand\`es and Y.~Plan.
\newblock Matrix completion with noise.
\newblock {\em Proceedings of the IEEE}, 9(6):925--936, 2010.

\bibitem[CR09]{RefWorks:390}
E.~J. Cand\`es and B.~Recht.
\newblock Exact matrix completion via convex optimization.
\newblock {\em Foundations of Computational Mathematics}, 9(6):717--772, 2009.

\bibitem[Dav03]{david2003chapter}
M.~David.
\newblock Chapter 20. an example inference task: clustering.
\newblock {\em Information Theory, Inference and Learning Algorithm}, 2003.

\bibitem[JMF99]{jain1999data}
A.~K. Jain, M.~N. Murty, and P.~J. Flynn.
\newblock Data clustering: a review.
\newblock {\em ACM computing surveys (CSUR)}, 31(3):264--323, 1999.

\bibitem[KMO10a]{RefWorks:698}
R.~Keshavan, A.~Montanari, and S.~Oh.
\newblock Matrix completion from a few entries.
\newblock {\em IEEE Transactions on Information Theory}, 56(6):2980--2998,
  2010.

\bibitem[KMO10b]{keshavan2010matrix}
R.~H. Keshavan, A.~Montanari, and S.~Oh.
\newblock Matrix completion from noisy entries.
\newblock {\em Journal of Machine Learning Research}, 11(Jul):2057--2078, 2010.

\bibitem[LS99]{lee1999learning}
D.~D. Lee and H.~S. Seung.
\newblock Learning the parts of objects by non-negative matrix factorization.
\newblock {\em Nature}, 401(6755):788, 1999.

\bibitem[LS01]{lee2001algorithms}
D.~D. Lee and H.~S. Seung.
\newblock Algorithms for non-negative matrix factorization.
\newblock In {\em Advances in neural information processing systems}, pages
  556--562, 2001.

\bibitem[RS05]{RefWorks:32}
J.~D.~M. Rennie and N.~Srebro.
\newblock Fast maximum margin matrix factorization for collaborative
  prediction.
\newblock In {\em Proc. 22nd int. conf. on Machine learning}, pages 713--719.
  ACM, 2005.

\bibitem[TS10]{tandon2010sparse}
R.~Tandon and S.~Sra.
\newblock Sparse nonnegative matrix approximation: new formulations and
  algorithms.
\newblock 2010.

\bibitem[XW05]{xu2005survey}
R.~Xu and D.~C. Wunsch.
\newblock Survey of clustering algorithms.
\newblock 2005.

\bibitem[XYWZ12]{MCNMF12}
Y.~Xu, W.~Yin, Z.~Wen, and Y.~Zhang.
\newblock An alternating direction algorithm for matrix completion with
  nonnegative factors.
\newblock {\em Frontiers of Mathematics in China}, 7(2):365--384, 2012.

\end{thebibliography}

%\bibliographystyle{IEEEbib}
%\bibliography{strings,refs}

\end{document}